# Defeasible Decisions: What the Proposal Is and Isn't

R. P. Loui
Dept. of Computer Science and
Dept. of Philosophy
Washington University
St. Louis, MO 63130
loui@ai.wustl.edu

## Abstract

In two recent papers, I have proposed a description of decision analysis that differs from the Bayesian picture painted by Savage, Jeffrey and other classic authors. Response to this view has been either overly enthusiastic or unduly pessimistic. In this paper I try to place the idea in its proper place, which must be somewhere in between.

Looking at decision analysis as defeasible reasoning produces a framework in which planning and decision theory can be integrated, but work on the details has barely begun. It also produces a framework in which the meta-decision regress can be stopped in a reasonable way, but it does not allow us to ignore meta-level decisions. The heuristics for producing arguments that I have presented are only supposed to be suggestive; but they are not open to the egregious errors about which some have worried. And though the idea is familiar to those who have studied heuristic search, it is somewhat richer because the control of dialectic is more interesting than the deepening of search.

## 1 What the Proposal Is.

"Defeasible Specification of Utilities" [Loui89a] and "Two Heuristic Functions for Decision" [Loui89b] proposed that decision analysis could profitably be conceived as defeasible reasoning. Analyzing a decision in one decision tree or model is an argument for doing a particular act. The result of analysis with a different tree of model is another argument. That there can be multiple arguments suggests that there can be better arguments and lesser arguments. Thus, arguments for decisions must be defeasible.

Response to this idea has been mixed, but often immoderate. This paper attempts to temper the reaction by saying what the proposal is and isn't.

### 1.1 Like Qualitative, Defeasible, Practical Reasoning.

The proposed defeasible reasoning about decisions is the natural extension of philosophers' defeasible practical reasoning about action. The difference is that our arguments for actions are quantitative, often invoking expected utility calculations.

In practical reasoning, reasoning about action is qualitative. If an act achieves a goal, that's a reason for performing that act. If an act achieves a goal but also invokes a penalty, and that penalty is more undesirable than the goal is desirable, that may be reason not to perform the act. Eveyone assumes that practical reasoning is defeasible in this way: that is, an argument for an action can be defeated by taking more into account in its deliberation. But this had never been formalized as defeasible reasoning, to my knowledge, because formalisms for defeasible reasoning are relatively new.

Now that we have such formalisms, we can write the general schemata for practical reasoning quite simply:

$(a)(d)$. $\ulcorner a$ ACHIEVES $d \wedge d$ IS-DES$\urcorner$ $\succ\!\!\!-$ $\ulcorner$DO $a\urcorner$,

$(a)(d)$. $\ulcorner a$ ACHIEVES $d \wedge d$ IS-UNDES$\urcorner$ $\succ\!\!\!-$
  $\ulcorner \neg ($DO $a)\urcorner$,

where " $\succ\!\!\!-$ " is a relation between sentences that corresponds roughly to our intuitive relation "is a reason for"; $(x)$ is a meta-language quantifier. Axioms that govern such a relation are described in [Loui87,88b] and [Simari89a]. They are similar to axioms given by other authors ([Geffner88,89], [Nute89], [Delgrande87]).

Reasons can be composed to form arguments. So prior to considering interference among contemplated actions, we might produce the argument:

"$a_1$ ACHIEVES $d_1 \wedge d_1$ IS-DES"  $\succ\!\!\!-$  "DO $a_1$"

"$a_2$ ACHIEVES $d_2 \wedge d_2$ IS-DES"  $\succ\!\!\!-$  "DO $a_2$"

"DO $a_1 \wedge$ DO $a_2$"  $\succ\!\!\!-$  "DO $a_1$ & $a_2$".

But there may be other arguments that disagree with this argument, such as

"$a_1$ & $a_2$ ACHIEVES $d_3 \wedge d_3$ IS-UNDES"  $\succ\!\!\!-$
  "$\neg($DO $a_1$ & $a_2)$".

In fact, we should be able to write our reasons in such a way that preference among arguments can be achieved with the specificity defeaters in defeasible inference, *i.e.*, those rules that tell us to prefer one argument over another if it uses more information.

Suppose I am reasoning about whether to rent an Alfa, though it incurs a big expense. An argument for renting an Alfa is based on the following reason:

"*rent-the-Alfa* ACHIEVES *drove-Alfa* $\wedge$ *drove-Alfa* IS-DES"  $\succ\!\!\!-$  "DO *rent-the-Alfa*".

A different argument which comes into conflict is based on the reason:

"*rent-the-Alfa* ACHIEVES *incurred-big-expense* $\wedge$ *incurred-big-expense* IS-UNDES"  $\succ\!\!\!-$
  "$\neg($DO *rent-the-Alfa*$)$".

There is no reason to choose among these arguments, so they interfere and neither justifies its conclusion. Suppose further that taking into account the desirability of driving the Alfa and the undesirability of incurring big expense, I judge the combination to be undesirable.

"*drove-Alfa* IS-DES ∧ *incurred-big-expense* IS-UNDES" >— "(*drove-Alfa & incurred-big-expense*) IS-UNDES".

Then there is a third argument, based on the combined reasons:

"*drove-Alfa* IS-DES ∧ *incurred-big-expense* IS-UNDES" >— "(*drove-Alfa & incurred-big-expense*) IS-UNDES"

"*rent-the-Alfa* ACHIEVES (*drove-Alfa & incurred-big-expense*) ∧ (*drove-Alfa & incurred-big-expense*) IS-UNDES" >— "¬(DO *rent-the-Alfa*)".

This argument disagrees with the first argument, which was in favor of renting the Alfa. But it takes into account all of the information that the first argument takes into account, and it does so in a way that cannot be counter-argued. So it is a superior argument; it defeats the first argument.

All of this reasoning about action is defeasible. There may be other arguments, based on what else we notice that renting the Alfa achieves, and what we may know about their desirability in various contexts. As more consequences of action are inferred, more arguments can be presented. Eventually, defeat relations among those arguments are proved. At any time, based on the pattern of defeat relations among presented arguments, there is either an undefeated justification for taking a particular action, or there are interfering arguments whose conflict has not been resolved. In the latter case, we might fall back on our un-tutored inclination (*e.g.*, to rent the Alfa). Sometimes we act for reasons; sometimes we act for very good reasons; sometimes we do not have the luxury of having unanimous reasons, or any reasons at all.

Of course, this qualitative practical reasoning is a very weak way of analyzing tradeoffs. It does not take into account known risks of actions, that is, known probabilities of acts achieving various effects.

## 1.2 But Quantitative and Risk-Sensitive.

What I have proposed is a quantitative version of this *defeasible reasoning* about decisions. An act achieves an effect with known probability, and we have independent reasons for the utilities of each of the resulting states. By weighing these independent utilities by their respective probabilities, we produce an argument for the utility of the act. With different independent reasons for the utilities of resulting states, we get different arguments. With different accounting of the possible results of an act, again, we produce different arguments. If we are clever, reasons can be written in an existing formalism for defeasible reasoning in such a way that those arguments that justify their conclusions are exactly those arguments that we would consider compelling among the multitude of potentially conflicting arguments.

Suppose I consider the possibility that my department will reimburse me for renting the Alfa, and calculate its probability to be 0.4. Based on expense and access to the Alfa, I assess the utilities of the various resulting states and calculate an expected utility for renting the Alfa. If it is greater than the utility of renting the econo-car, it represents an argument for renting the Alfa.

$u(dept\text{-}pays;\ rent\text{-}the\text{-}Alfa\ \text{BASED-ON}\ expense;\ whether\text{-}drove\text{-}Alfa) = 10$ utils

$u(\neg dept\text{-}pays;\ rent\text{-}the\text{-}Alfa\ \text{BASED-ON}\ expense;\ whether\text{-}drove\text{-}Alfa) = -1$ utils

$Expected\ u(rent\text{-}the\text{-}Alfa\ \text{BASED-ON}\ expense;\ whether\text{-}drove\text{-}Alfa;\ whether\text{-}dept\text{-}pays) = 3.4$ utils

$u(rent\text{-}econo\text{-}car) = 2$ utils

therefore, defeasibly, *rent-the-Alfa*.

But if instead I consider expense, access to the Alfa, and the dissatisfaction of my department chairman, in the assessment of utilities, then I produce a different argument.

$u(dept\text{-}pays;\ rent\text{-}the\text{-}Alfa\ \text{BASED-ON}\ expense;\ whether\text{-}drove\text{-}Alfa;\ how\text{-}chairman\text{-}reacts) = 8$ utils

$u(\neg dept\text{-}pays;\ rent\text{-}the\text{-}Alfa\ \text{BASED-ON}\ expense;\ whether\text{-}drove\text{-}Alfa;\ how\text{-}chairman\text{-}reacts) = -4$ utils

$Expected\ u(rent\text{-}the\text{-}Alfa\ \text{BASED-ON}\ expense;\ whether\text{-}drove\text{-}Alfa;\ how\text{-}chairman\text{-}reacts;\ whether\text{-}dept\text{-}pays) = 0.8$ utils

$u(rent\text{-}econo\text{-}car) = 2$ utils

therefore, defeasibly, *rent-econo-car*.

As before, there may be other arguments, based on other contingencies to be analyzed (*e.g.*, whether I can fool the accounting secretary, whether it rains, etc.) and other factors that affect the independent reasons for valuing various states of the world (*e.g.*, how my colleagues react, how my friends react, etc.). Normally, in decision analysis we require all such reasons to be taken into account in advance. For computational and foundational reasons, this is not so here.

One way this reasoning can be formalized is with the following axiom schemata, which presume as default that there is a linear-additive structure to utility when exceptions are not known.

Properties, such as $P$ and $Q$, make basic contributions to the utility of a state in which they are known to hold. If the contribution of $P$ is $x$ and the contribution of $Q$ is $y$, that provides a reason for taking the contribution of the conjunction to be the sum of $x$ and $y$.

Ax.1. $(x)(y)(P)(Q).\ \ulcorner contr(P) = x\ \&\ contr(Q) = y \urcorner$ >— $\ulcorner contr(P\ \&\ Q) = x + y \urcorner$.

This is defeated if we know independently the contribution of $P$ & $Q$ to be something other than the sum of the individual contributions.

"contr" maps properties to utility contributions. Any information about this mapping, together with the knowledge that property $P$ holds in state $s$, provides a reason for taking the utility of $s$ to be the contribution of $P$. If $P$ & $Q$ is known to hold in $s$, then taking $u(s)$ to be the contribution of $P$ & $Q$ will result in a better argument for what is the utility of $s$.

Ax.2. $(P)(s)$. $\ulcorner T(P,s) \urcorner \succ\!\!- \ulcorner u(s) = contr(P) \urcorner$.

Here, $T(P,s)$ says that $P$ holds in $s$.

Finally, if event $E$ is known to have probability $k$ in state $s$, then this provides a reason for taking the utility of $s$ to be the weighted sum of the children's utilities.

Ax.3. $(E)(s)(k)$. $\ulcorner T(prob(E)=k,s)\urcorner \vdash \ulcorner u(s) = u(<E;s>)k + u(<\neg E;s>)(1-k) \urcorner$.

## 1.3 Better Detailed Than Some Have Thought.

Ax.1 has been criticized on two counts: first, for permitting a utility pump (anonymous referee), and second, for assuming independence of contributions([Thomason89]). The first criticism is wrong if we add the obvious requirement that

Ax.i. $(P)(Q) \vdash \ulcorner (P \equiv Q) \supset contr(P) = contr(Q) \urcorner$.

Then $contr(P_1) = 10$ does not provide reason for $contr(P_1 \& P_1) = 20$; the axioms governing the construction of arguments require that they be consistent. The second criticism is right, but empty: indeed, we assume independence of contributions. But that is a defeasible assumption, and when it is an incorrect assumption, we expect that it is made known as an explicit exception. The properties *does-smoke* and *has-cancer* typically co-occur. The individual contribution of *does-smoke* in a state might be $-20$, and the individual contribution of *has-cancer* in a state might be $-50$, but the joint contribution of does-smoke and has-cancer might be an exception to additivity, $-60$. This exception must be stated explicitly.

Figures 1 through 5 show some arguments for utility valuations, and interations. In each case, the utility of state $s$ is at issue. Here I am assuming Simari's system [Simari89a].

Figure 1 shows the most basic argument for the utility of $s$ based on the contribution of $P$, which holds in $s$. The conclusion, $u(s) = 5$, is based on the theory below it. It rides over a bold horizontal line whenever it is justified. The theory consists of a set of defeasible rules, which are depicted as connected digraphs (the convention here is that arrows always go up). Sources must be given as evidence. Contingent sentences used in these graphs are underlined and are important in determining specificity of various arguments. sentences below the vertical line are setences given as evidence that are used to produce the conclusion, but not to activate any defeasible rules.

Figure 2 shows a similar argument, where two defeasible rules are used instead of one. Multiple in-directed edges are understood as representing conjunction.

Figure 3 shows two arguments that conflict. When there is defeat of one argument by another, the defeating argument has a large arrowhead. The defeating argument here is the most basic argument that uses the expected utility axiom (Ax. 3). Defeat is clear to see here because

$T(P \& R, E \mid s) \& T(P \& R, E \mid s) \vdash T(P,s)$.

Figure 4 shows conflict among two arguments that determine the contributions of properties through defeasible arguments.

Figure 5 shows an argument that uses both expected utility and defeasible reasoning about the contributions of properties. Since the contingent sentences are those underlined, specificity holds. Note that if the sentences concerning *contr* had been regarded as contingent, there would not have been specificity.

## 1.4 One Way To Integrate Planning and Decision Theory.

The original motivation for the use of non-monotonic language was to find a concise specification of utility functions mapping descriptions of the world into the reals, when descriptions are collections of sentences in a first order predicate logical language. There is no way for planning research to exploit the existing ideas for decision-making under known risk if there is no practical way to represent the relative desirability of descriptions of the world.

Concise representation is not trivial. If there are $2n$ logicaly independent atomic formulae that contribute to the valuation of states, then there are $2^n$ sentences in the Lindenbaum-Tarski algebra of atomic sentences. Descriptions of the world number $3^n$ if we allow only conjunction of atomic formulae, or $2^{(2^n)}$ if we allow arbitrary disjunction. A brute force method would have us specify the utilities of each of the $2^n$, at the very least.

Decision theory avoids the burden of this specification by presuming that the world has only a small number of relevant properties, expressed by propositions or atomic formulae. It is significant that Jeffrey's Logic of Decision [Jeffrey65] never explicitly considers algebras larger than $2^{10}$; even then, it is considered in a case in which there is indifference among all of the descriptions.

Planning avoids the burden of specification by making no graded distinctions among descriptions. Descriptions of the world either satisfy all of the goals or they do not. In this case, there is a short description of the utility mapping, on the order of the number of goals, and independent of the size of the descriptions.

When utilities are real-valued, instead of $0-1$, there may still be concise descriptions of the utility mapping. We presume that there is a way to construct utilities of states from the structure of their descrip-tions. That is, there are regularities in the utility mapping that can be expressed in a small number of rules, and possibly some exceptions to those rules. One kind of regularity that can be exploited is separability of the contributions of various parts of the description. The natural way to partition a description is by its atomic sentences. This is the regularity that axiom Ax.1 represents: a linear-additive separability reminiscent of multi-attribute decompositions of utility.

Now we can pose decision-problems on worlds whose descriptions are as complicated as they are in planning problems.

There is plenty of work to be done here. The next question to be studied is how to interpret the planning ideas of goal-directed search. In our present representation of utility, there is no clear line between goals and non-goals. There are sentences that contribute heavily to the utility of a world's description, and sentences that are relatively inconsequential, all things being equal. This can depend on context, on what else is contained in the description. It is unclear how to intepret goal-directed chaining not only because of the conception of goals, but also because of the nature of chaining. Since the world of decision-making under risk is not deterministic, chaining may have to distinguish events that are and are not under the direct control of the agent.

One possible use of planning's backward-chaining from goals in the current framework is as follows. Suppose on the confrontation of a decision problem that there is some way of identifying a small number of saliently desirable and undesirable states. For the moment, treat all events as controllable, *i.e.*, as if they could be chosen, even if they have a low probability of occurring. Now backward chain to identify some of the sequences of acts and events that lead to these salient states. These are not exhaustive. They do not represent the likely outcomes of sequences of actions. But they are starting points for constructing arguments.

Consider each path. Arguments are constructed for the value of the path's probability and the value of the terminal state's utility. These become subarguments for the argument to take (or avoid) the actions contained in the path. Each path defines a decision tree exaggerated in depth and narrowness. Of course, there will be disagreements: some lines of envisionment correspond to optimism, some to pessimism. It is a simple matter to combine the trees into a single argument to resolve the disagreements. The result is a fairly bushy decision tree. A good argument should also consider what states are achieved with the bulk of the probability.

This is just one interpretation of existing AI planning practice on the sequencing of actions. There may be others to be investigated.

## 1.5 One Way to Avoid Small Worlds.

The use of defeasible reasoning is intended to avoid the problem of small worlds (see also [Edwards89] and [D'Ambrosio&Fehling89]). Savage worried briefly about how much detail there ought to be in decision models. There are limitations on what we can explicitly consider in models. Could it be that taking more into account explicitly can alter the choice of act? On Savage's view, this is impossible. On my view, it is a fact of life. Savage's solution was to retreat to an ideal in which all considerations not explicit are implicit; revealing them cannot alter the decision. My solution is to accept non-monotonicity in deliberation, and accept that the model can be improved. The answer is not to suppose that all small worlds are as good as big worlds, but to posit that the smaller worlds can be avoided by enlarging worlds over time.

This opens my deliberators to hypothetical dutch books. An agent could commit to an act, and it could be the case that further deliberation would have recommended an act incompatible with the first: incompatible in the sense that both acts taken simultaneously would guarantee loss of money, if not of utility. But as has been known for years now, such hypothetical dutch books are inocuous. No one would take the second action after committing to implement the first as well [Kyburg78]. In fact, our model of deliberation does not even include simultaneous acts.

## 1.6 One Way to Avoid Computing Relevance.

The problem of small worlds is one problem for resource-limited rational reasoners. The proposed view of decision-making copes with two other notorious problems of limited rationality. It is worth considering the ways in which it copes.

Spectators of AI in psychology and philosophy [Pylyshyn87] have discussed two related problems: first, how to compute what is relevant without looking at the irrelevant, *i.e.*, how to know which stones not to turn without turning them; second, how to think that enough thinking has been done, so that it is time to act. Some have misinterpreted these as the frame problem, but they are better called the problem of turning stones, and Hamlet's problem, respectively.

Both problems involve a regress. In the case of Hamlet's problem, the regress is that meta-reasoning seems required optimally to stop the reasoning, and meta-meta-reasoning is required optimally to stop the meta-reasoning, and so on. This proposal avoids the regress by accepting sub-optimality at the highest level, or better, by questioning the notion of optimality in infinitely extendable analyses. We do not preclude meta-reasoning (as below in section 2) but also do not require it.

It is less important for the model to be optimal at a given instant when it will be altered immediately in the next instant. Models can still be good or bad, wonderful or awful, and meta-reasoning may be required to choose one over the other, and meta-meta-reasoning to improve the choice. But at some level, one has to take one's chances.

In the case of turning stones, the regress is one of deciding what is relevant to a problem, which might require looking at irrelevant aspects of the problem.

Moreover, the decision procedure itself might be irrelevant. This proposal avoids the regress by not requiring that it get started. Most aspects of the problem that are irrelevant do not find their way into the analysis. By hypothesis, many relevant aspects also do not find their way into the analysis, because deliberation time expires: an argument is discovered that establishes that an act should be performed now. Conversely, some irrelevancies actually do find their way into the analysis. Good search and chaining strategies avoid irrelevancies. Good control strategies avoid prematurity of action or hopeless postponement of action. Distinguishing good and bad strategies is a computation taken off-line.

## 2  What the Proposal Isn't.

Some have thought the proposal to be something that it is not.

### 2.1  Just Game Tree Search.

The proposal is a lot like the proposal to do heuristic search on a game tree. It is true that the idea will be unsurprising to those who have studied heuristic search [Pearl88, Hansson&Mayer89]. Apparently, all one does is substitute expected utility for the minimax criterion. The proposal differs in more substantive ways, however.

The obvious difference is that there are two ways to supercede a heuristic evaluation of a node here. One way is to deepen the analysis and use the children's expected utility for the heuristic evaluation. Another way is simply to involve more properties that hold at that node in the heuristic evaluation.

There is a sense in which taking a node's children into account just is taking more of its properties into account. Looked at that way, there is only one way to improve a heuristic evaluation, namely, to take more properties into account. Still, on this view, this is a generalization of heuristic search, where heuristic evaluation is improved in a computation only by deepening.

The more important difference from game tree search is less obvious. It has to do with the aggregation of arguments and defeat relations among them, to produce justification for an action. The dialectic of argument, counter-argument, defeat, and reinstatement is more complex than the simple deepening of heuristic search. Consider a situation where there is an argument to do a particular act, and a disagreeing argument, which says to do a different act. There is a third argument, too, which counterargues and defeats the second argument, thereby reinstating the first argument. These are shown in figure 6 and recounted below:

Argument 1:
$T(P, a_1) \succ\!\!- u(a_1) = contr(P)$
$T(Q, a_2) \succ\!\!- u(a_2) = contr(Q)$
$contr(P) = 3$
$contr(Q) = 1$
$u(a_1) > u(a_2) \succ\!\!- \text{DO}(a_1)$

This is an argument to do $a_1$ based on a comparison with $a_2$.

Argument 2:
$T(P, a_1) \succ\!\!- u(a_1) = contr(P)$
$T(R, a_3) \succ\!\!- u(a_3) = contr(R)$
$contr(P) = 3$
$contr(Q) = 5$
$u(a_1) < u(a_3) \succ\!\!- \neg(\text{DO}(a_1))$.

This is an argument that disagrees with the first argument.

Argument 3:
$T(R \,\&\, S, a_3) \succ\!\!- u(a_3) = contr(R \,\&\, S)$
$contr(R \,\&\, S) = 2$.

This argument counterargues argument 2 at the point where argument 2 contends that $u(a_3) = 5$. It does not augment argument 1 in any way, but it defeats argument 2, thus reinstating argument 1.

The third argument could conceivably be integrated into the first argument to produce a more comprehensive argument: an argument that defeats the second argument by itself. But that would be somewhat complex and is not required in the presence of the second argument's rebuttal. Why build into an argument a defense to every possible objection, when each objection can be rebutted as it arises?

I do not think there are game tree search situations that correspond to this state of defeasible deliberation on decision.

### 2.2  Excluding Meta-Level Analysis.

Prominent work on limited rationality is being done on decision-theoretic meta-reasoning (esp. [Horvitz88], [Russell and Wefald89], [Etzioni89]). This proposal seems to conflict with their approaches because it does not require meta-reasoning. But it does not preclude meta-reasoning, and sometimes such reasoning is useful. I do not have notation for representing nor axiom schemata for generating reasons and arguments of the following kind, but conceivably they could be produced. The first have to do with meta-reasoning that controls the attention at the object level:

$(a_1 \succ a_2 \text{ in } M_1) \wedge (M_1 \text{ is not worth expanding})$ is reason to do $a_1$ now

or

$(a_1 \succ a_2 \text{ in } M_1) \wedge (do\text{-}now(a_1) \succ expand(M_1))$
$\succ\!\!- do\text{-}now(a_1)$.

These would be reasons that say that there is no net perceived value of expanding the model. As the cited authors have pointed out, the preference to expand a model may be based on an expected utility computation. I would add that computation of such a utility will be defeasible:

$u(do\text{-}now(a_1)) = 15 \wedge u(expand(M_1)) = 10$
$\vdash do\text{-}now(a_1) \succ expand(M_1)$

$T(intended\text{-}act\text{-}succeeds, do\text{-}now(a_1)) \wedge contr(intended\text{-}act\text{-}succeeds) = 15 \succ\!\!- u(do\text{-}now(a_1)) = 15$

$T(prob(find\text{-}better\text{-}act) = .3, expand(M_1))$, etc.

As an aside, my discussions have not been about arguments for acting "now" as oppsed to acting "later." I have presumed that time simply expires, leaving an apparent best act at the moment. To produce arguments for action "now" would seem to re-open Hamlet's problem: given such an argument to act "now," do we take the time to seek a counter-argument?

The second kind of meta-reasoning has to do with arbitrating among disagreeing arguments at the object level, when no defeat relation is known to hold:

$(a_1 \succ a_2$ in Arg1 $\wedge$

$a_2 \succ a_1$ in Arg2 $\wedge$

Arg1 is based on short-term-considerations $\wedge$

Arg2 is based on long-term-considerations) $\succ\!\!-\ a_1 \succ a_2$

So I do not see meta-reasoning as incompatible with this proposal. We have so little experience with mechanizing simple arguments at the object level, however, that the focus of attention remains there.

## 2.3 One Particular Heuristic Function.

This proposal does not live or die with the multi-attribute suggestion for representing utility concisely. A second heuristic is exhibited in [Loui89b] based on [Schubert88]. In order to have utility expectations, all that is needed is some representation of utility on sentential descriptions of the world. Practical necessity demands that there be some regularity that can be exploited for compact representation.

In order to achieve defeasibility in our deliberation about decisions, all that are needed are independent reasons for valuing a state based on lists of properties that can be proved to hold in that state. Those lists of properties are not complete, and reasons for valuations based on incomplete properties need not bear relation to valuations based on probabilities of the omitted properties.

In fact, I expect that heuristics for utility will vary from individual to individual, and will depend on application.

## 2.4 Terrible Computation.

A valid concern [Pearl89] is that I am substituting something whose effective computation is well understood (heuristic search) with something whose effective computation has yet to be achieved (a variety of non-monotonic reasoning). This is true to the extent that deliberation on decision is just heuristic search, and special cases of defeasible reasoning do not yield to special purpose, effective inference procedures. If the only dialectic envisioned is the succession of arguments based on successively deepened trees, all of which defeat their predecessors, then this defeasible reasoning is analogous to heuristic search. And it can be implemented without much ado. But defeasible reasoning about decision can be more interesting than that. Until we discover patterns of dialectic for decision that lead to special algorithms, we are stuck with the general framework for defeasible reasoning. This situation is not so bad: *dialectic in defeasible reasoning has good prospects for being controlled reasonably well under resource limitation.*

## 2.5 Necessarily Quantitative.

Reasons need not be quantitative. Consider

$a_1$ is better than the usual risk is reason to do $a_1$

or

$a_1$ achieves my aspirations in this context is reason to do $a_1$.

Qualitative reasons make especially good sense at the meta-level:

(the difference between $a_1$ and $a_2$ is small) and ($a_1$ is robust) is reason to do $a_1$.

A reason suggested by Doyle as a tie-breaker is:

can't choose between $a_1$ and $a_2$ is reason to do $a_1$.

Again, I have no schemata for generating these kinds of reasons, and no way to weigh arguments based on these reasons against arguments based on quantitative considerations. But I believe that a full theory of deliberation would include them, or be able to reduce them to quantitative reasons.

## 2.6 Complete.

Finally, it should be admitted clearly that this proposal is not complete. The integration of planning techniques, the exploration of meta-reasoning and qualitative reasons, the production of reasons for acting now, control of search and dialectic, and experience with particular heuristics are all things to be done. All we can do at present is produce expected utility arguments at various levels of detail. We do, however, have a PROLOG-based implementation of the underlying defeasible reasoning system [Simari89b] and see no major obstacle in using the schemata Ax.1 – Ax.3 with some help unifying terms within functions.

# 3 An Open Conversation with Raiffa.

There is a device through which to take the measure of this proposal's break from Bayesian tradition, and at the same time to see the inutitiveness of what is being proposed. Consider the following hypothetical conversation with the great decision theorist, Howard Raiffa. I phone him at his Harvard office to solicit his best decision analysis under resource limitation.

Ron: I'm at the San Francisco airport. I have this decision problem – whether to rent an Alfa. Can you help?

Raiffa: Sure. I have this theory, you know. What are all the relevant distinctions among states? All the effects of events? All the available courses of action?

Ron: You want me to list them all? I don't have time! Am I paying for this phone call?

Raiffa: Yes, I see. Hmm. Ok. Confine your attention to the important ones.

Ron: How important?

At this point, there are two good responses.

Raiffa 1: Well, let's make a model of the expected utility of omitting various considerations.

Raiffa 2: Well, let's just start and see what comes to mind and refine the model later.

The Bayesians want to think that the first answer is the only legitimate one. Meanwhile, it is the second answer that makes sense to us. What is the logic of decision analysis based on this second answer? This is the question that I have been attempting to answer.

## 4 References.

Figure 1.

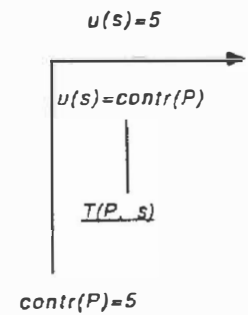

Figure 2.

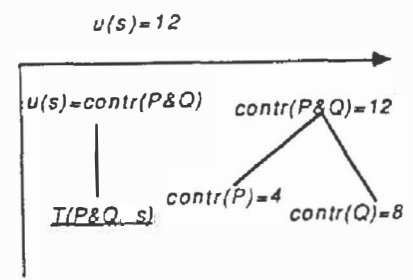

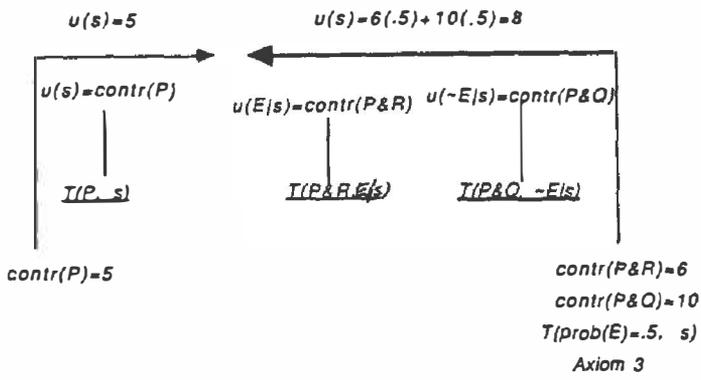

Figure 3.

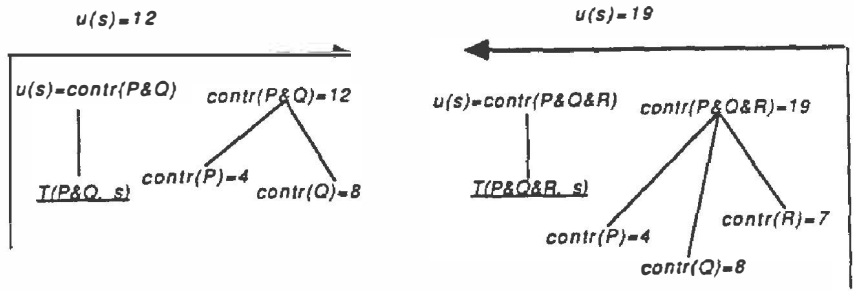

Figure 4.

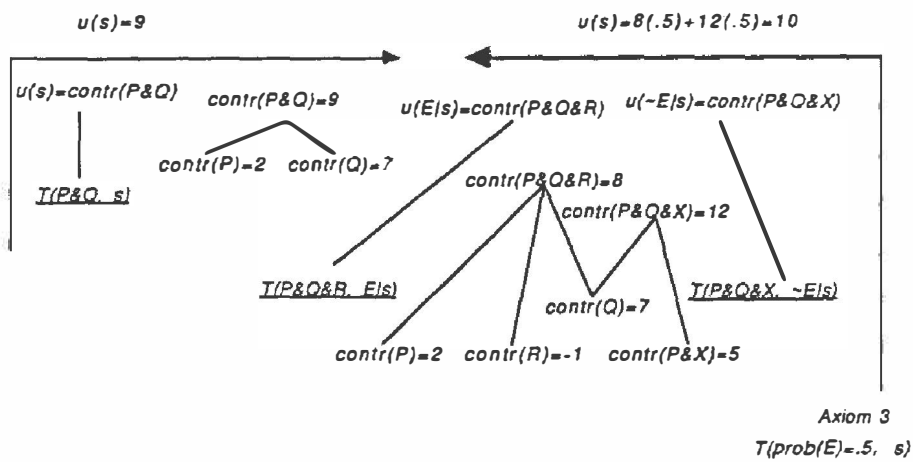

Figure 5.

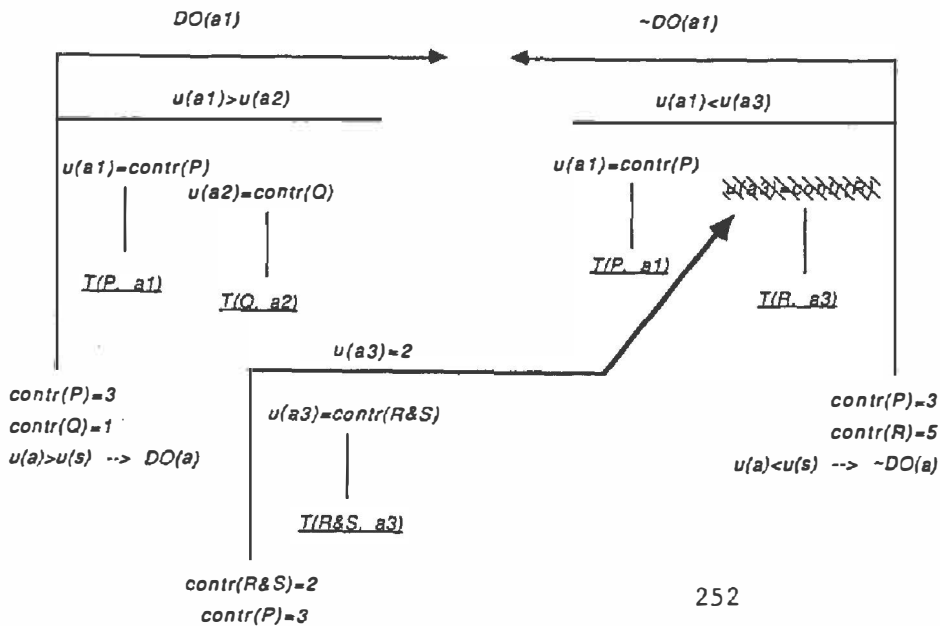

Figure 6.